\documentclass[conference]{IEEEtran}
\IEEEoverridecommandlockouts
\usepackage{cite}
\usepackage{amsmath,amssymb,amsfonts}
\usepackage{algorithm} 
\usepackage{algorithmic}
\usepackage{graphicx}
\usepackage{textcomp}
\usepackage{xcolor}

\usepackage{booktabs}
\usepackage{tabu}
\usepackage{diagbox}
\usepackage{bm}
\usepackage{tabularx, array, bbding}
\usepackage{caption2}
\usepackage{multirow}
\def\BibTeX{{\rm B\kern-.05em{\sc i\kern-.025em b}\kern-.08em
    T\kern-.1667em\lower.7ex\hbox{E}\kern-.125emX}}
\begin{document}

\title{PoissonSeg: Semi-Supervised Few-Shot Medical Image Segmentation via Poisson Learning\\
\thanks{* Corresponding author: Jianwei Lu, Ye Luo.}
}

\author{\IEEEauthorblockN{1\textsuperscript{st} Xiaoang Shen}
\IEEEauthorblockA{\textit{School of Software Engineering} \\
\textit{Tongji University}\\
Shanghai, China \\
sxa@tongji.edu.cn}
\and
\IEEEauthorblockN{2\textsuperscript{nd} Guokai Zhang}
\IEEEauthorblockA{\textit{School of Optical-Electrical and Computer Engineering} \\
\textit{University of Shanghai for Science and Technology}\\
Shanghai, China \\
zhangguokai\_01@163.com}
\and
\IEEEauthorblockN{3\textsuperscript{rd} Huilin Lai}
\IEEEauthorblockA{\textit{School of Software Engineering} \\
\textit{Tongji University}\\
Shanghai, China \\
lhl@tongji.edu.cn}
\and
\IEEEauthorblockN{4\textsuperscript{rd} Jihao Luo}
\IEEEauthorblockA{\textit{School of Computing} \\
\textit{National University of Singapore}\\
Singapore \\
E0576164@u.nus.edu}
\and
\IEEEauthorblockN{5\textsuperscript{rd} Jianwei Lu$^*$}
\IEEEauthorblockA{\textit{School of Software Engineering} \\
\textit{Tongji University}\\
Shanghai, China \\
jwlu33@tongji.edu.cn}
\and
\IEEEauthorblockN{6\textsuperscript{rd} Ye Luo$^*$}
\IEEEauthorblockA{\textit{School of Software Engineering} \\
\textit{Tongji University}\\
Shanghai, China \\
yeluo@tongji.edu.cn}
}
\maketitle
\begin{abstract}
The application of deep learning to medical image segmentation has been hampered due to the lack of abundant pixel-level annotated data. Few-shot Semantic Segmentation (FSS) is a promising strategy for breaking the deadlock. However, a high-performing FSS model still requires sufficient pixel-level annotated classes for training to avoid overfitting, which leads to its performance bottleneck in medical image segmentation due to the unmet  need for annotations. Thus, semi-supervised FSS for medical images is accordingly proposed to utilize unlabeled data for further performance improvement. 
Nevertheless, existing semi-supervised FSS methods has two obvious defects: (1) neglecting the relationship between the labeled and unlabeled data; (2) using unlabeled data directly for end-to-end training leads to degenerated representation learning. To address these problems, we propose a novel semi-supervised FSS framework for medical image segmentation. The proposed framework employs Poisson learning for modeling data relationship and propagating supervision signals, and Spatial Consistency Calibration for encouraging the model to learn more coherent representations. In this process, unlabeled samples do not involve in end-to-end training, but provide supervisory information for query image segmentation through graph-based learning. We conduct extensive experiments on three medical image segmentation datasets (i.e. ISIC skin lesion segmentation, abdominal organs segmentation for MRI and abdominal organs segmentation for CT) to demonstrate the state-of-the-art performance and broad applicability of the proposed framework.
\end{abstract}


\section{Introduction}
The automatic segmentation from medical image is of clinically significance and has been widely used in disease diagnosis, treatment and prognosis. Notably, deep learning based segmentation models have achieved satisfying performance due to their remarkable feature extraction ability. However, training a fully-supervised deep learning model usually requires abundant labeled data, which is extremely challenging in clinical practice due to the cumbersome annotation work.

To tackle this challenge, few-shot learning that trains a model and makes predictions for an unseen class (known as query) under the guidance of the knowledge learned from a few labeled samples (known as support) has been accordingly proposed \cite{RaviS}\cite{FinnC}. Nevertheless, when applied to medical images, few-shot semantic segmentation has not gained promising results as training a FSS model needs a large dataset with many pixel-level annotated classes to avoid overfitting. And medical image annotation is more troublesome and expensive compared to that of natural images, to say nothing of pixel-level annotation of various classes (e.g. lesions, organs, and other human body parts). Though abundant pixel-level annotated images are unavailable, the images with only class labels are comparatively sufficient. Therefore, how to fully explore the supervisory information from these image-level annotated images (denoted as auxiliary images and also referred to as unlabeled data/images/samples below for lacking pixel-level labels) to train a high-performing FSS model for medical image segmentation remains a core challenge. 
The emergence of semi-supervised FSS offers a specific solution. Recently, some works \cite{Feyjie}\cite{Abdel} apply semi-supervised FSS to medical images, but these methods are faced with performance bottlenecks, which can be attributed to: (1) ignoring the relationship between the labeled and unlabeled samples; (2) using unlabeled samples directly for end-to-end training (e.g., adopting consistency loss \cite{Feyjie} and augmenting data with pseudo-labels\cite{Abdel} ), which misleads models to learn the degenerated representations.

\begin{figure}[!t]
\centering
\includegraphics[width = 9cm]{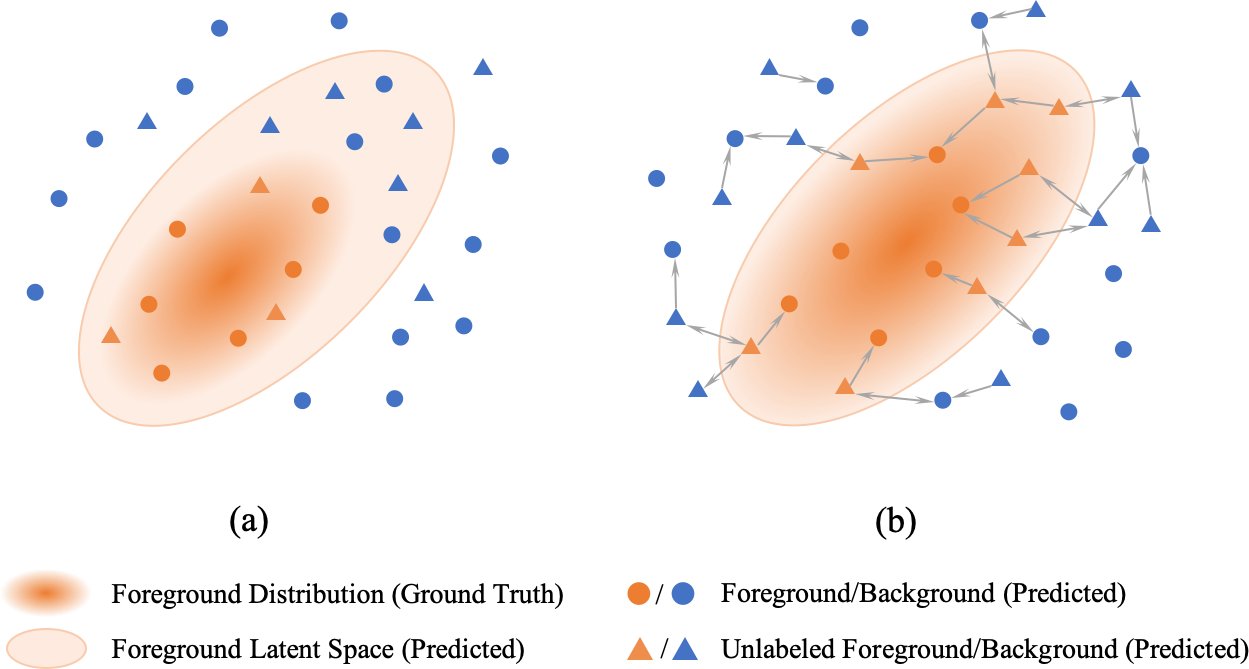}
\caption{An example of the foreground distribution with (right) and without (left) Poisson learning. The orange/blue circles represent foreground/background local prototypes of the support images, while the orange/blue triangles represent foreground/background local prototypes of the auxiliary images. And the gray arrows represent the communication between two samples in label inference (only a small fraction are displayed).}
\label{ssfs_intro}
\end{figure}

To address aforementioned problems, in this paper, we propose a novel semi-supervised FSS framework named as PoissonSeg. Typically, our proposed PoissonSeg network exploits semi-supervised learning with two modules, i.e., Poisson Learning module and Spatial Consistency Calibration (SCC) module. The Poisson Learning module builds a graph to model relationship between labeled and unlabeled images, and then conducts label inference for information passing. Specifically, we construct graphs with local prototypes rather than global prototypes or pixel-level feature vectors, since previous works \cite{YangB}\cite{YuQ}\cite{Ouyang} verify that multiple local prototypes can enrich semantic representations and avoid semantic ambiguity.
Fig.\ref{ssfs_intro} illustrates how the introduction of Poisson learning assists the classification for foreground/background local prototypes: in Fig.\ref{ssfs_intro} (a), the whole foreground latent space cannot be covered by the distribution of foreground local prototypes, since there lacks effective channels for unlabeled samples to communicate supervisory information with labeled samples. Meanwhile, in Fig.\ref{ssfs_intro} (b), when applying Poisson learning, graphs are constructed for data samples to conduct label inference, and thus the foreground distribution and the foreground latent space are more consistent. Besides, we propose the SCC module to maintain the spatial consistency \cite{RohB} and further improve the model performance. The SCC module encourages the pixels of the same class to be consistent by propagating the transformed features of the similar pixels. In this way, the model can learn more coherent representations and yield segmentation results with considerable spatial smoothness. On the whole, although the unlabeled samples do not involve in end-to-end training, they indeed provide supervision information for query image segmentation through graph-based learning.

Overall, the main contributions of this paper can be summarized as:

1) We propose PoissonSeg, the first semi-supervised FSS framework that adopts graph-based learning to model the relationship between labeled and unlabeled images, to the best of our knowledge. It fully exploits the supervisory information from unlabeled images and provides effective supervision for query image segmentation.

2) We propose spatial consistency calibration to help model learn coherent representations and thus maintain the spatial smoothness of the segmentation results. When combined with Poisson learning, SCC can significantly boost the performance of our model.
 
3) Extensive experiments on multiple medical image segmentation datasets demonstrate the effectiveness and broad applicability of our PoissonSeg framework. Our work also highlights the importance of modeling the relationship between unlabeled and labeled data in semi-supervised FSS and provides new ideas for future works on medical image segmentation.

\section{Related Work}

\subsection{Few-shot semantic segmentation}

Few-shot learning was initially introduced to address classification problems with very limited annotations for each class. However, placing general semantic segmentation in a few-shot scenario is much more challenging, since dense pixel prediction for new classes needs to be performed with only a few support samples. Inspired by \cite{Snell}, existing works on few-shot segmentation typically adopt a metric-based strategy, where the network usually measures the similarity (e.g. cosine similarity) between the pixel feature vectors from the query images and the prototypes from support images to produce segmentation results. OSLSM \cite{Shaban} firstly introduces a two-branch network consisting of a support branch and a query branch for FSS, and many other works also follow this two-branch architecture design \cite{ZhangC,Nguyen,WangK}. Zhang et al. \cite{ZhangC} propose to utilize graphs and attention mechanism to model structured segmentation data. Nguyen et al. \cite{Nguyen} present an apporach that encourages high feature activations on the foreground and low feature activations on the background. In PANet \cite{WangK}, discriminative embedding prototypes are obtained by applying prototype alignment regularization, which promotes the consistency of prototypes through exchanging the roles of support and query samples. More recently, such prototype-based methods catch more attention and got fully developed \cite{YangB,WangH,Azad,LiuB}. For example, Yang et al. \cite{YangB} point out the semantic ambiguity problem caused by using a single prototype, and they improve the prototype representation learning by correlating diverse image regions with multiple prototypes to solve this problem. Wang et al. \cite{WangH} leverage a probabilistic latent variable model to infer the distribution of the prototype, which enhances model's generalization ability to handle the inherent uncertainty and the intra-class variations. 

The success of few-shot segmentation on natural images drives more and more researchers to apply it to medical image processing. Roy et al. \cite{Roy} integrate ‘squeeze \& excite’ blocks to the network for the segmentation of volumetric medical images with only a few annotated slices. Sun et al. \cite{SunL} develop an global correlation module to capture the correlation between a support and query image and incorporate it into a global correlation network, which is proved effective on both abdomen MRI and CT segmentation tasks. Ouyang et al. \cite{Ouyang} design an adaptive local prototype pooling module to overcome foreground-background imbalance problem. And superpixel-based pseudo-labels are generated to offer self-supervision during training.

\begin{figure*}[h]
\centering
\includegraphics[width=\textwidth]{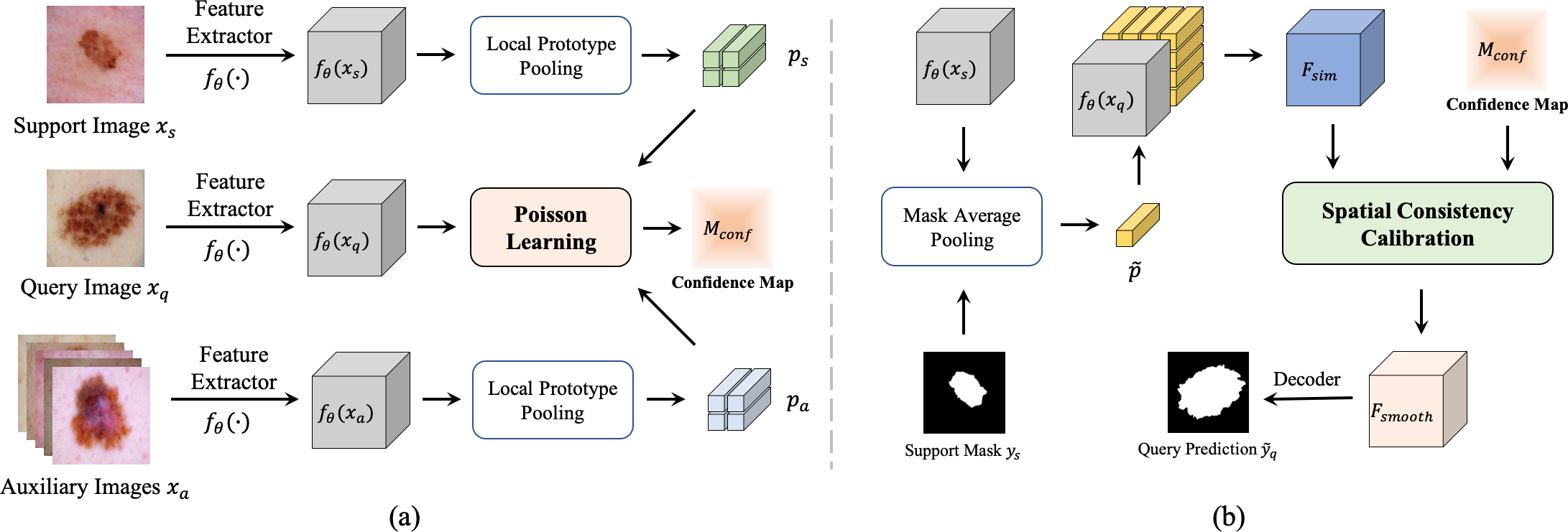}
\caption{An overview of our PoissonSeg Network. (a) The main network for Poisson learning to generate the confidence map; (b) The network of the inference mainly by the spatial consistency calibration module.}
\label{network_main}
\end{figure*}

\subsection{Semi-supervised learning for semantic segmentation}

Due to the scarcity of pixel-level annotations, semi-supervised learning, which addresses the classification problem by incorporating extra unlabeled data, has been applied to semantic segmentation tasks. Generally, consistency loss and weak annotations are the two main ideas used in semi-supervised semantic segmentation problem. Consistency loss usually forces the model to make consistent predictions on the unlabeled data under different transformations or minimizes the discrepancy between labeled and unlabeled data over some properties. Feyjie et al. \cite{Feyjie} propose to include surrogate tasks that learn a mapping between noised images and their original counterparts for semi-supervised few-shot medical image segmentation. Li et al. \cite{LiS} enforce the consistency of the geometric shape constraint on the labeled and unlabeled set to regularize the model learning. Instead of focusing on data-level consistency, Luo et al. \cite{LuoX} build dual-task consistency regularization by jointly predicting  a pixel-wise segmentation map and a geometry-aware level set segmentation map of the target for both labeled and unlabeled data. Moreover, various forms of weak annotations (e.g. image labels, bounding boxes and scribbles) are also used as supervisory signals for model's training on unlabeled data. Zhang et al. \cite{ZhangW} propose an Expectation-Maximization (EM) algorithm to estimate segmentation labels for the weakly annotated images (only with image-level or bounding box annotation). Lin et al. \cite{LinD} design a graph-based model that propagates information from scribbles to unmarked pixels and learns network parameters. 

\section{Methodology}

In this section, we first give the problem formulation for the semi-supervised FSS on medical images. The details of our network architecture are introduced with a focus on the Poisson learning and the spatial consistency calibration module. Finally, we present the loss function and the end-to-end training procedure of our model. 

\subsection{Problem Formulation}\label{AA}
In semi-supervised FSS, besides a support set $\mathcal{D}_{s}=\{(x_s,y_s)\}$ and a query set $\mathcal{D}_q=\{(x_q,y_q)\}$ that both contain pixel-level annotated images of different classes, an auxiliary set $\mathcal{D}_a=\{x_a\}$ consisting of images with only image-level annotations is also provided. Models are trained on training classes $\mathcal{C}_{train}$ and tested on testing classes $\mathcal{C}_{test}$ ($\mathcal{C}_{train} \cap \mathcal{C}_{test}=\varnothing$). In training stage, given support images $\mathcal{S}=\{(x_s,y_s)\}_{i=1}^{K \times C} \subset \mathcal{D}_{s}$ and auxiliary images $\mathcal{A}=\{x_a\}_{i=1}^{M \times C} \subset \mathcal{D}_{a}$, the segmentation model takes $\mathcal{S}$ and $\mathcal{A}$ as reference and then makes pixel-level predictions for every query image in $\mathcal{Q}=\{(x_q,y_q)\}_{i=1}^{N_q} \subset \mathcal{D}_{q}$. Here, $x$ and $y$ represent the image and its associated pixel-level annotated mask. $C$ denotes the number of classes. $K$ and $M$ are the numbers of the images for each class in set $\mathcal{S}$ and $\mathcal{A}$, respectively. Such a training process involving $(\mathcal{S},\mathcal{A},\mathcal{Q})$ comprises a training episode \cite{Shaban}, which is also defined as a $C$-way $K$-shot semi-supervised few-shot segmentation problem. The case is quite the same with the testing stage. Note that the auxiliary set is also provided for testing.

\subsection{Network Architecture}
\label{sec_net}
The proposed PoissonSeg network architecture is illustrated in Fig.\ref{network_main}. Our network consists of two parts: (a) the Poisson learning part for the generation of confidence map; (b)  inference part that combines the information from $\mathcal{S}$ and $\mathcal{A}$ through spatial consistency calibration and yields the final segmentation results.

As shown in Fig.\ref{network_main}-(a), the support images $x_s$, auxiliary images $x_a$ and query images $x_q$ pass through a weight-sharing feature extractor $f_{\theta}(\cdot)$ (we choose VGG-16\cite{VGG} as the backbone.) and produce their corresponding feature maps $f_{\theta}(x_s) \in \mathbb{R}^{C \times H \times W}$, $f_{\theta}(x_a) \in \mathbb{R}^{C \times H \times W}$ and $f_{\theta}(x_q) \in \mathbb{R}^{C \times H \times W}$. Then we apply local prototype pooling to $f_{\theta}(x_s)$ and $f_{\theta}(x_a)$ to enrich the semantic representations of prototypes. Here the local prototype pooling process can be formulated as:
\begin{equation}
    p_{i,j}=\phi_{h,w}(f_{\theta}(x))(i,j), 
\end{equation}
where $\phi_{h,w}(\cdot)$ denotes average pooling with a pooling window size of $(h,w)$ and  $p_{i,j}$ is the obtained local prototype at location $(i,j)$. We denote the local prototype extracted from $f_{\theta}(x_s)$ along with its corresponding class label as $(p_{s},\hat{y}_p)$, and view the feature vector at each location of feature maps $f_{\theta}(x_q)$ as an individual feature sample $v_q$. Thus we can further define $\mathcal{P}=\{(p_s,\hat{y}_p)\} \bigcup \{p_a\} \bigcup \{v_q\}$, where the labels of $\{v_q\}$ can be determined by Poisson learning, a graph-based semi-supervised learning algorithm on $\mathcal{P}$. Afterwards, the confidence map $M_{conf}$ can be constructed using the predicted labels of $\{v_q\}$. Details of Poisson learning can be referred to Sec.\ref{sec_poisson}.

In the inference part, as shown in Fig.\ref{network_main}-(b), mask average pooling is first applied to $f_{\theta}(x_s)$ to get prototype:
\begin{equation}
    \tilde{p}=\frac{\underset{i}{\sum}\underset{j}{\sum}\tilde{y}_{i,j}f_{\theta}(x_s)(i,j)}{\underset{i}{\sum}\underset{j}{\sum}\tilde{y}_{i,j}},
\end{equation}
where $\tilde{y}_{i,j} \in \mathbb{R}^{H \times W}$ is the down-sampled version of $y_s$. And $\tilde{p}$ is further expanded and convolved with $f_{\theta}(x_q)$ to produce similarity maps $F_{sim}$. Then the prior knowledge of confidence map is incorporated to similarity maps by the SCC module. The SCC module is designed to encourage the model to learn coherent representations thus ensure the spatial smoothness of prediction. The calibrated similarity maps $F_{smooth}$ are finally decoded to yield final prediction of the query image. Details of the SCC module can be referred to Sec.\ref{sec_scc}. 

\subsection{Poisson Learning }
\label{sec_poisson}
Poisson learning is initially proposed to tackle graph-based semi-supervised learning problems and is proved effective for our semi-supervised FSS task as well. We redefine $\mathcal{P}=\{p_s\} \bigcup \{p_a\} \bigcup \{v_q\}=\{p_1,p_2,...,p_n\}$ ($n=n_s+n_a+n_q$), where $\{p_i\}_{i=1}^{n_s}=\{p_s\}$, $\{p_i\}_{i=n_s+1}^{n_s+n_a}=\{p_a\}$ and $\{p_i\}_{i=n_s+n_a+1}^{n}=\{p_q\}$. Note that the first $n_s$ vertices have the ground truth label, denoted as $\{\hat{y}_1,\hat{y}_2,...,\hat{y}_{n_s}\}$. Hence, we can build a graph with the feature vectors in $\mathcal{P}$ as vertices and denote the edge weight between $p_i$ and $p_j$ as $w_{ij}$. Here we define $w_{ij}$ as:
\begin{equation}
    w_{ij}=exp(-4|p_i-p_j|^2/d_K(p_i)^2),
\end{equation}
where $d_K(p_i)$ is the distance between $p_i$ and its $K$-th nearest neighbor. We assume that $w_{ij}=w_{ji}$ and $w_{ij}\ge 0$. The degree of a specific vertice $p_i$ is defined as $d_i=\sum_{j=1}^{n}w_{ij}$. Furthermore, we define the weight matrix as $W=[w_{ij}]$, the degree matrix as $D=diag(d_i)$. Accordingly, the unnormalized graph Laplacian is $L=D-W$. Let $\overline{y}=\frac{1}{n_s}\sum_{i=1}^{n_s}\hat{y}_i$ be the average label vector and let $\varrho_{ij}=1$ if $i=j$ and $\varrho_{ij}=0$ if $i \neq j$. Here, we want to learn a classifier $g:\mathcal{P}\rightarrow\mathbb{R}^k$  ($k$ is the number of classes and $k=2$ for foreground/background classification) by solving the Poisson equation:
\begin{equation}
    Lg(p_i)=\sum_{j=1}^{n_s}(\hat{y}_j-\overline{y})\varrho_{ij}\quad\text{for $i=1,2,...,n$},
\label{l_eq}
\end{equation}
which satisfies $\sum_{i=1}^{n}d_ig(p_i)=0$.

To solve Eq.(\ref{l_eq}), some notations are introduced first. Let $Y=[\hat{y}_i] \in \mathbb{R}^{k \times n_s}$ denote the label matrix of $\{p_s\}$ and $\overline{Y}=[\overline{y}] \in \mathbb{R}^{k \times n_s}$ denote the expanded matrix of $\overline{y}$. Let $Q=[Y-\overline{Y},\textbf{0}^{k \times (n - n_s)}]$ denote the initial label matrix of all data. Moreover, we initialize $R$, the prediction label matrix of $\mathcal{P}$, as $\textbf{0}^{n \times k}$. Thus we can get a stable $g(\cdot)$ with $T$ iterations using following equation:
\begin{equation}
    R_{t+1}=R_{t}+D^{-1}(Q^T-LR_{t}),
\end{equation}
where the subscript $t$ is the iteration index.
After obtaining $R_T$, we build a query label matrix $U=R_T[:,n-n_q:n]$ using the last $n_q$ columns of $R_T$ and  reshape it into $U_{r} \in  \mathbb{R}^{k \times H \times W}$. Finally, we apply soft-max to $U_r$ and select the last channel as the confidence map: 
\begin{equation}
    M_{conf} = \psi(U_r)[k,:,:],
\end{equation}
where $\psi(\cdot)$ denotes the soft-max operation. 

\subsection{Spatial Consistency Calibration }
\label{sec_scc}
In Sec.\ref{sec_net}, the prototype $\tilde{p}$ is expanded and convolved with $f_{\theta}(x_q)$ to produce similarity maps, which can be described as:
\begin{equation}
    F_{sim}=Conv(f_{\theta}(x_q) \parallel \varphi_1(\tilde{p})),
\end{equation}
where $\parallel$ represents the concatenation operator, $\varphi_1:\mathbb{R}^{C}\rightarrow\mathbb{R}^{C \times H \times W}$ denotes the expanding function and $Conv(\cdot)$ is the convolution operation. We incorporate the prior knowledge of $M_{conf}$ into $F_{sim}$ mainly by the element-wise multiplication:
\begin{equation}
    F_{fuse} =F_{sim} \odot \varphi_2(M_{conf}),
\end{equation}
where $\odot$ represents element-wise multiplication operator and $\varphi_2:\mathbb{R}^{H \times W}\rightarrow\mathbb{R}^{C \times H \times W}$ denotes the expanding function. $F_{fuse}$ now contains all information from both the similarity maps $F_{sim}$ and the confidence map $M_{conf}$. 

The confidence map generated by Poisson learning encodes the prior knowledge about the possible label of each pixel of a query image. However, the potential drawback of the confidence map $M_{conf}$ generated in Sec.\ref{sec_poisson} is that we take the feature vector of each pixel as an individual feature sample while totally neglecting the spatial context in feature maps. Therefore, directly fusing the similarity maps with the confidence map could lead to severe spatial inconsistency and thus unsmooth segmentation results. To get coherent representations, for each pixel feature vector $v_i$ in $F_{fuse}$, we compute its 
consistency calibration $\tilde{v}_i$ by re-weighting the linear transformation of all pixel feature vectors $v_j$ in $F_{fuse}$ with the similarity between $v_i$ and $v_j$ and averaging them:
\begin{equation}
\label{ssc_eq}
    \tilde{v}_i = \frac{1}{HW}\sum_{j}^{HW}ReLU(\sigma(v_i,v_j)) \cdot h(v_j),
\end{equation}
where $\sigma(\cdot)$ is a similarity function (e.g. cosine similarity), $ReLU(\cdot)$ denotes the ReLU activation function and $h(\cdot)$ denotes the linear transformation function (here we choose two-layer MLP.). We denote the smoothness calibrated similarity map as $F_{smooth}$, which is then decoded to produce final segmentation results.

\subsection{Semi-supervised Learning}
The training process of our proposed PoissonSeg network consists of two phases: (1) pretraining without Poisson learning; (2) online training with complete network architecture. In the first phase, we pretrain our network on $(\mathcal{S},\mathcal{Q})$ and do not employ Poisson learning in order to train the feature extractor $f_{\theta}(\cdot)$ producing discriminative features. In the second phase, we perform end-to-end online training on $(\mathcal{S}, \mathcal{A}, \mathcal{Q})$ using the complete network design. 
Dice loss is adopted for network training and can be defined as:
\begin{equation}
    \mathcal{L}_{\text {Dice }}=1-\frac{2|X \cap Y|+\epsilon}{|X|+|Y|+\epsilon}.
\end{equation}
Here, X and Y represents the predicted mask and the ground truth mask separately. $\epsilon$ acts as a smoothing factor that controls numerical stability. 
\begin{table*}[!h]
\setlength{\belowcaptionskip}{6pt}
\fontsize{8}{12}\selectfont
\caption{Comparisons the effectiveness of the different components of our model on ISIC Dataset.}
\label{tab1}
\centering
\setlength{\tabcolsep}{4mm}{
\begin{tabular}{l|l|l|l|l|l|l|l|l|l}
\hline
Method & AK & BCC & BK & DF & MEL & NV & SCC & VASC & Mean \\
\hline
\hline
Baseline  & 43.66 & 42.90 & 28.50 & 58.62 & 59.34 & 27.15 & 46.37 & 36.86 & 42.93\\
Baseline + PL  & 55.36 & 51.01 & 41.28 & 63.29 & 71.43 & 50.95 & 59.72 & 57.67 & 56.34\\
Baseline + SCC  & 48.95 & 43.18 & 29.29 & 60.32 & 62.63 & 33.15 & 54.84 & 40.37 & 46.59\\
\textcolor{black}{\textbf{Proposed}} & \textcolor{black}{\textbf{59.87}} & \textcolor{black}{\textbf{52.04}} & \textcolor{black}{\textbf{48.35}} & \textcolor{black}{\textbf{67.19}} & \textcolor{black}{\textbf{73.24}} & \textcolor{black}{\textbf{58.70}} & \textcolor{black}{\textbf{67.51}} & \textcolor{black}{\textbf{68.13}} & \textcolor{black}{\textbf{61.88}}\\
\hline
\end{tabular}}
\end{table*}

\section{Experiments}
\subsection{Dataset}
To demonstrate the effectiveness and general applicability of our model for different semi-supervised few-shot medical segmentation tasks, the evaluations is performed on three datasets: ISIC skin lesion segmentation, abdominal organs segmentation for MRI and abdominal organs segmentation for CT. Considering the scarcity of pixel-level annotated images in clinical practice, for each dataset, only a small fraction of data is utilized to build $\mathcal{D}_s$ and $\mathcal{D}_q$ while the remainder is taken as $\mathcal{D}_a$. And all our experiments are conducted under the 1-way 1-shot scenario, which is similar to \cite{Ouyang}. 

\textbf{ISIC Dataset} Two skin lesion dermoscopic datasets (i.e. ISIC-2017 and ISIC-2019) are used to investigate the capability of our model for the few-shot segmentation of color medical images. ISIC-2017 \cite{Codella} and ISIC-2019 datasets are both provided by the International Skin Imaging Collaboration (ISIC). The ISIC-2017 dataset provides 2000 pixel-level annotated dermoscopic images (i.e. 374 melanoma images, 254 seborrheic keratosis images and 1372 nevus images) for skin lesion segmentation. The ISIC-2019 dataset contains 25,331 images across 8 different classes (i.e. AK, BCC, BKL, DF, MEL, NV, SCC and VASC) with aforementioned 3 classes in ISIC-2017 included. However, these images only have class labels and lack pixel-level annotations. These two datasets are merged as our ISIC dataset by eliminating the overlapping classes from ISIC-2019. Then, for each class, we randomly select 15 images as fully annotated data (each image has both the mask and class label.) and 45 images as auxiliary data (each image only has its class label.). Since there are some images belonging to the five classes in ISIC-2019 lacking masks, we manually annotate these images under the guidance of several experienced dermatologists.

\textbf{Abdomen-MRI Dataset}\label{abd-mri} This dataset is provided by ISBI 2019 Combined Healthy Abdominal Organ Segmentation Challenge \cite{Kavur} and consists of 20 3D T2-SPIR MRI scans. We separate out three quarters of the slices to build $\mathcal{D}_a$.

\textbf{Abdomen-CT Dataset}\label{abd-ct} This dataset is provided by MICCAI 2015 Multi-Atlas Abdomen Labeling Challenge \cite{Landman} and contains 30 3D abdominal CT scans. Also, We separate out three quarters of the slices as $\mathcal{D}_a$.

For the ISIC Dataset, the label set contains eight classes, i.e., actinic keratosis (AK), basal cell carcinoma (BCC), benign keratosis (BK), dermatofibroma (DF), melanoma (MEL), Nevus (NV), squamous cell carcinoma (SCC) and vascular (VSAC). For the Abdomen-MRI/CT Dataset, we choose liver, left kidney (LK), right kidney (RK) and spleen to constitute a shared label set. And we apply the same data pre-processing pipeline (e.g. extracting 2D slice from the 3D sequence, etc.) as \cite{Ouyang} did to the Abdomen-MRI/CT Dataset.

\subsection{Experiment Set-up}
\subsubsection{Evaluation Metric} Dice Similarity Coefficient (DSC) score, commonly used in medical image segmentation researches, is employed to evaluate the performance of our model. The definition of DSC score is formulated as:
\begin{equation}
    \text { DSC }=1-\frac{2|A \cap B|}{|A|+|B|},
\end{equation}
where A and B are the predicted mask and the ground truth, respectively. 

\subsubsection{Implementation Details}
Our model is implemented with Pytorch and trained on a NVIDIA GTX 1080Ti GPU for 100 epochs. The pretraining phase takes about 40 epochs and the other 60 epochs is for online training. We use SGD as optimizer while set the learning rate to 0.001 and the momentum to 0.9. All the experiments follow a standard five-fold cross-validation procedure.

\subsection{Ablation Study}
\begin{figure}[h]
\centering
\includegraphics[width=8.8cm]{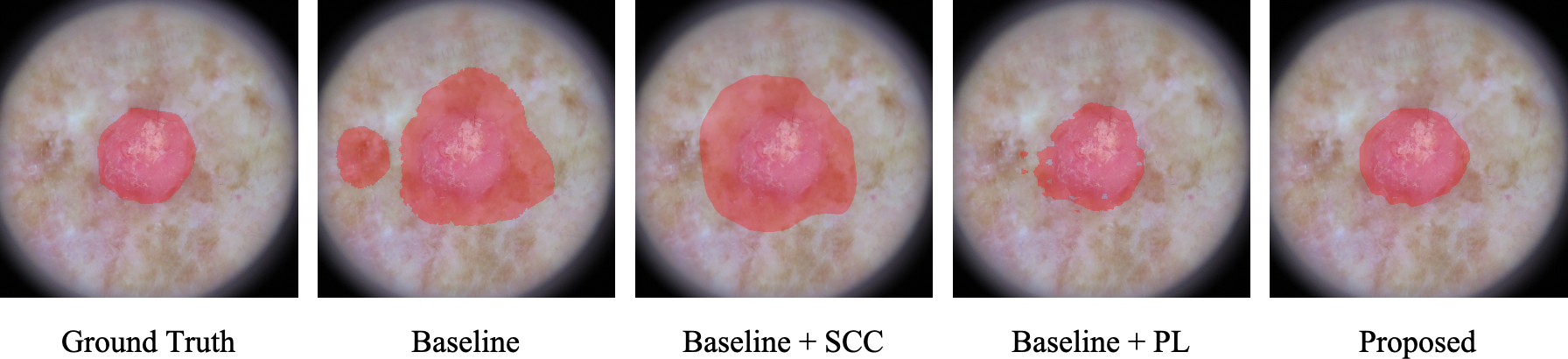}
\caption{Visualization the effectiveness of different components of the proposed method on ISIC Dataset.}
\label{isic_fig}
\end{figure}

Following experiments are conducted on ISIC Dataset to verify the effectiveness of each component in our network. We remove the Poisson learning module as well as the SCC module (i.e. Eq.(\ref{ssc_eq})) denoted as 'Baseline'. 'Baseline+PL' represents adding the Poisson learning module to the 'Baseline' without adopting SCC, while 'Baseline+SCC' retains the SCC module but without Poisson learning. And 'Proposed' is our complete design that incorporates both of these two modules. The ablation study results are shown in Table \ref{tab1}. Significant performance improvement can be seen by adding Poisson learning to the 'Baseline'. But applying only SCC brings comparatively slight performance gain. When combining both Poisson learning and SCC, our method can achieve the best results. To study how each component contributes to the segmentation task, we display the qualitative results in Fig.\ref{isic_fig}. It can be intuitively observed that 'Baseline+PL' can make more precise prediction than 'Baseline' at the expense of spatial smoothness. However, with the addition of SCC module, the spatial inconsistency problem can be solved and the model can produce more smooth segmentation results.

\subsection{Comparison with State-of-the-arts}
\begin{table*}[!htbp]
\setlength{\belowcaptionskip}{6pt}
\fontsize{8}{12}\selectfont
\caption{Segmentation result comparisons among the state-of-the-arts on ISIC Dataset.}
\label{tab2}
\centering
\setlength{\tabcolsep}{4mm}{
\begin{tabular}{l|l|l|l|l|l|l|l|l|l}
\hline
Model & AK & BCC & BK & DF & MEL & NV & SCC & VASC & Mean \\
\hline
\hline
PANet\cite{WangK}  & 41.46 & 46.98 & 29.73 & 57.82 & 56.58 & 38.18 & 49.07 & 47.17 & 45.87\\
SENet\cite{Roy}  & 42.06 & 38.75 & 32.03 & 59.54 & 53.04 & 31.80 & 50.55 & 35.86 & 42.95\\
GCN-DE\cite{SunL}  & 49.19 & 46.32 & 38.69 & 55.40 & 63.27 & 46.87 & 57.80 & 60.46 & 52.25\\
ASGNet\cite{LiG} & 47.88 & 48.24 & 36.15 & 62.37 & 58.79 & 42.51 & 54.36 & 49.05 & 49.92\\
\hline
SSFLNet\cite{Feyjie}  & 38.58 & 45.60 & 31.97 & 46.52 & 45.72 & 24.14 & 41.27 & 43.17 & 39.62\\
ALPNet\cite{Ouyang}  & 54.75 & 47.98 & 40.02 & 62.63 & 61.06 & 51.42 & 63.46 & 58.45 & 54.97\\
\textcolor{black}{\textbf{PoissonSeg}} & \textcolor{black}{\textbf{59.87}} & \textcolor{black}{\textbf{52.04}} & \textcolor{black}{\textbf{48.35}} & \textcolor{black}{\textbf{67.19}} & \textcolor{black}{\textbf{73.24}} & \textcolor{black}{\textbf{58.70}} & \textcolor{black}{\textbf{67.51}} & \textcolor{black}{\textbf{68.13}} & \textcolor{black}{\textbf{61.88}}\\
\hline
\end{tabular}}
\end{table*}

\begin{table*}[!htbp]
\setlength{\belowcaptionskip}{6pt}
\fontsize{8}{12}\selectfont
\caption{Segmentation result comparisons among the state-of-the-arts on Abdomen-MRI/CT Dataset.}
\label{tab3}
\centering
\setlength{\tabcolsep}{4mm}{
\begin{tabular}{l|l|l|l|l|l|l|l|l|l|l}
\hline
\multirow{2}{*}{Model} & \multicolumn{5}{c|}{Abdomen-MRI} & \multicolumn{5}{c}{Abdomen-CT} \\
\cline{2-11}
~ & Liver & RK & LK & Spleen & Mean & Liver & RK & LK & Spleen & Mean\\
\hline
\hline
PANet\cite{WangK} & 33.62 & 23.44 & 21.75 & 27.58 & 26.60 & 29.12 & 18.15 & 18.44 & 22.68 & 22.10\\
SENet\cite{Roy}  & 28.84 & 27.17 & 23.50 & 29.38 & 27.22 & 27.17 & 23.75 & 24.68 & 21.79 & 24.35\\
GCN-DE\cite{SunL}  & 38.36 & 34.12 & 29.49 & 28.23 & 32.55 & 34.00 & 26.83 & 28.29 & 24.21 & 28.33\\
ASGNet\cite{LiG} & 41.03 & 30.48 & 27.65 & 30.21 & 32.34 & 32.77 & 19.42 & 22.07 & 26.97 & 25.31\\
\hline
SSFLNet\cite{Feyjie}  & 24.51 & 15.74 & 18.83 & 22.37 & 20.36 & 21.25 & 14.03 & 13.52 & 18.37 & 16.79\\
ALPNet\cite{Ouyang}  & 53.86 & 52.44 & 47.70 & 49.44 & 50.86 & 54.06 & 42.39 & 44.66 & 49.98 & 47.77\\
\textcolor{black}{\textbf{PoissonSeg}}& \textcolor{black}{\textbf{61.03}} & \textcolor{black}{\textbf{53.57}} & \textcolor{black}{\textbf{50.58}} & \textcolor{black}{\textbf{52.85}} & \textcolor{black}{\textbf{54.51}} & \textcolor{black}{\textbf{58.74}} & \textcolor{black}{\textbf{47.02}} & \textcolor{black}{\textbf{50.11}} & \textcolor{black}{\textbf{52.33}} & \textcolor{black}{\textbf{52.05}}\\
\hline
\end{tabular}}
\end{table*}

\begin{figure*}[!htbp]
\centering
\includegraphics[width=\textwidth]{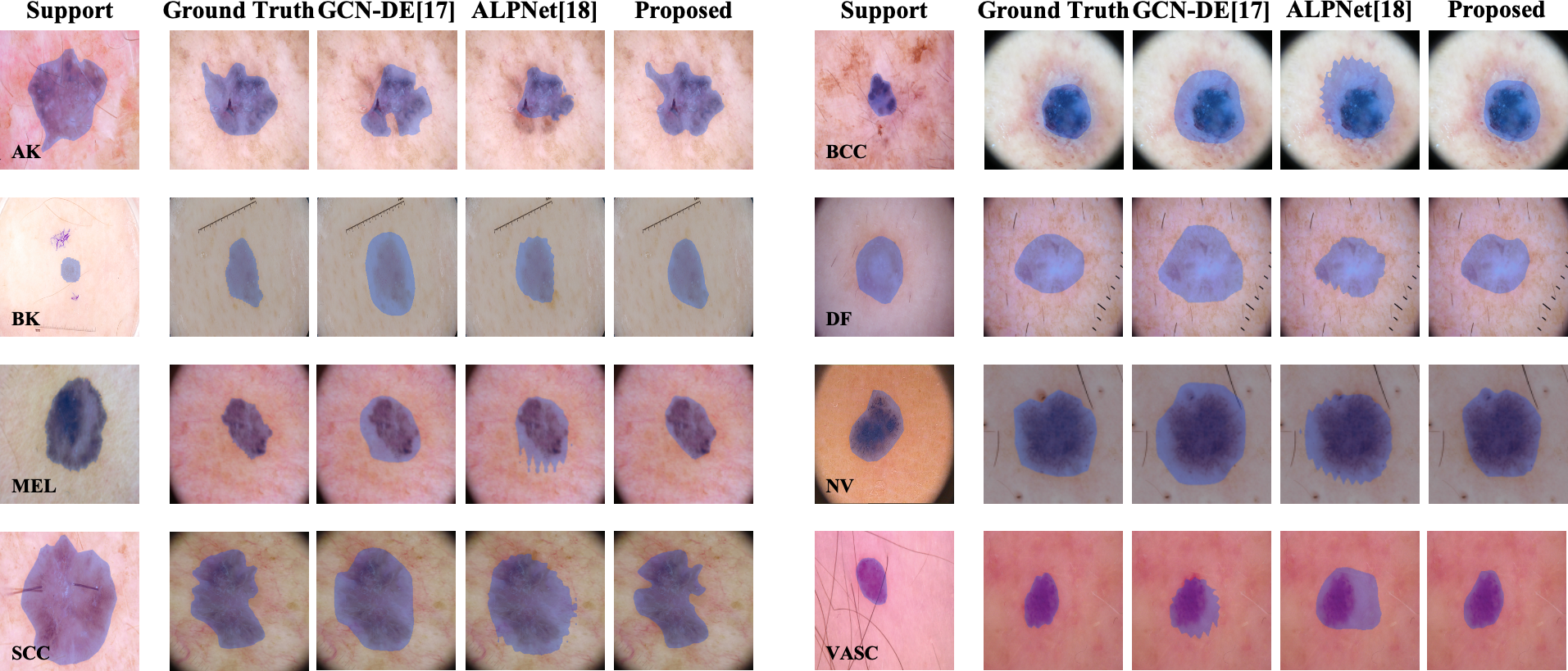}
\caption{Visualization of the comparisons between our method and the other two models on ISIC Dataset.}
\label{visual_1_fig}
\end{figure*}

\begin{figure*}[!htbp]
\centering
\includegraphics[width=\textwidth]{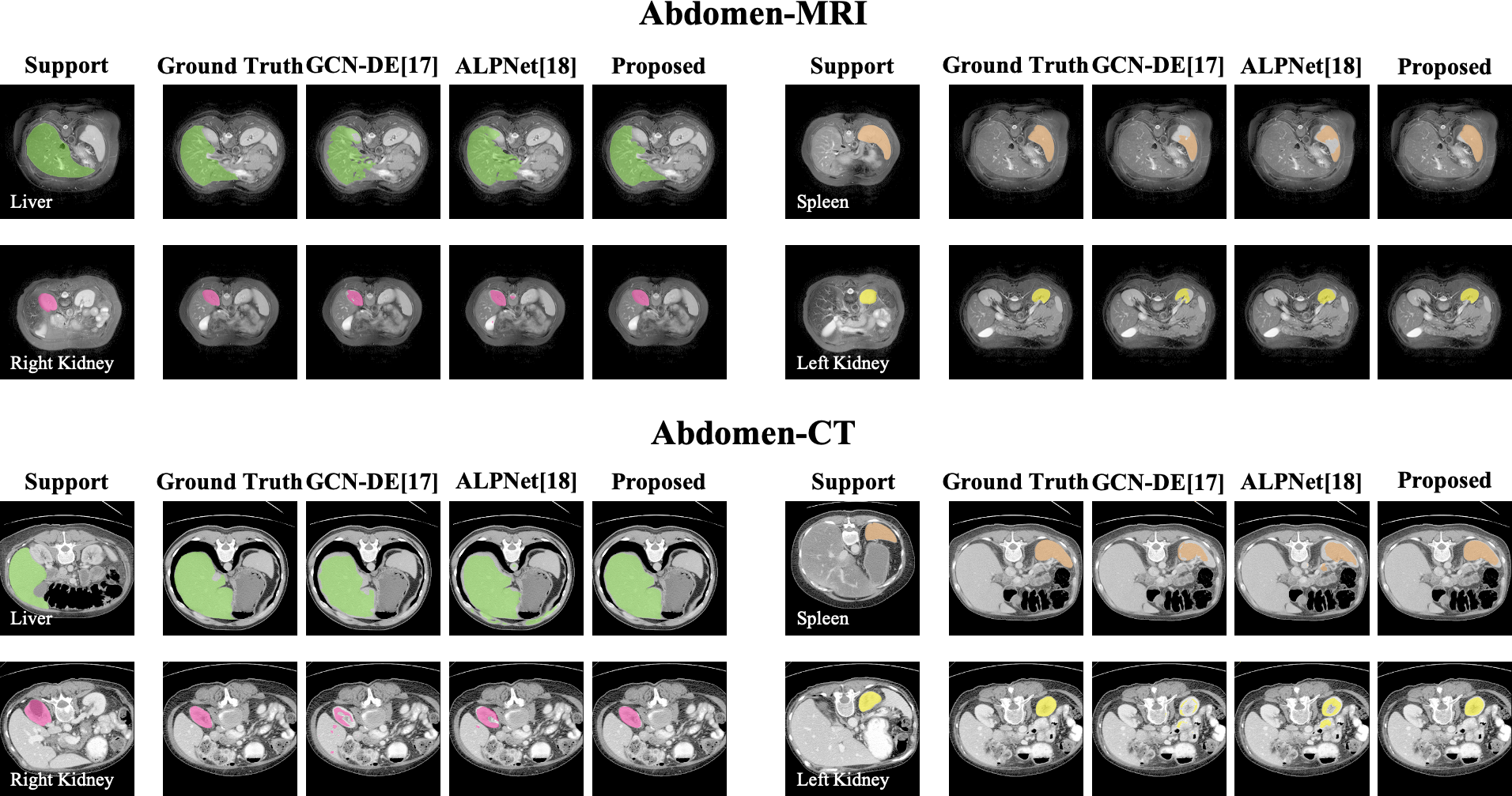}
\caption{Visualization of comparisons between our model and the other two models on Abdomen-MRI/CT Dataset.}
\label{visual_2_fig}
\end{figure*}

Table \ref{tab2} and Table \ref{tab3} show the comparison results of our model with other state-of-the-art methods on ISIC Dataset and Abdomen-MRI/CT Dataset respectively. Specifically, here we choose both fully-supervised FSS methods (PANet\cite{WangK}, SENet\cite{Roy}, GCN-DE\cite{SunL} and ASGNet\cite{LiG}) and semi/self-supervised ones (SSFLNet \cite{Feyjie} and ALPNet \cite{Ouyang}) for comprehensive comparisons. For fairness, in these implementations, we use exactly the same feature extraction backbone (i.e. VGG-16\cite{VGG}) as in PoissonSeg. The comparison result indicates that our PoissonSeg could outperform other methods on both ISIC Dataset and Abdomen-MRI/CT Dataset. Furthermore, the qualitative results of different methods are illustrated in Fig.\ref{visual_1_fig} and Fig.\ref{visual_2_fig}. Significantly, here we only select one model with the best performance (i.e. GCN-DE\cite{SunL} and ALPNet \cite{Ouyang}) from the fully-supervised FSS methods and the semi/self-supervised FSS methods respectively to compare with our method in Table \ref{tab2} and Table \ref{tab3}. From these two figures, we can observe that our PoissonSeg can locate the target region more precisely and make fewer false position predictions than other methods. Moreover, the segmentation results yielded by our method have smoother contour and are more similar to the ground truth. Overall, PoissonSeg produces more satisfying segmentation results on targets of different modalities with various color, shape and size.

\section{Conclusion}
In this work, we propose a novel semi-supervised  few-shot medical image segmentation framework named as PoissonSeg. Aiming at the limitations of annotated data scarcity in clinical practice, PoissonSeg exploit semi-supervised FSS with Poisson learning and Spatail Consistency Calibration. Specifically, Poisson learning constructs graphs to model relationship between labeled and unlabeled samples, and propagates supervision information via label inference. And SCC maintains the spatial consistency and encourages coherent representation learning. Extensive  experiments on skin lesion images and anatomical abdomen CT/MRI images validate the state-of-the-art performance and broad applicability of our model. In the future, we will extend our PoissonSeg framework to the segmentation of  volumetric medical images.

\end{document}